\documentclass[10pt, final, conference, letterpaper, twoside, twocolumn]{IEEEtran}
\IEEEoverridecommandlockouts
\usepackage[utf8]{inputenc}
\usepackage{cite}
\usepackage{amsmath,amssymb,amsfonts}
\usepackage{algorithmic}
\usepackage{graphicx}
\usepackage{textcomp}
\usepackage{xcolor}
\def\BibTeX{{\rm B\kern-.05em{\sc i\kern-.025em b}\kern-.08em
    T\kern-.1667em\lower.7ex\hbox{E}\kern-.125emX}}
\usepackage{balance}
\usepackage[hidelinks]{hyperref}
\usepackage{tikz}
\usepackage[caption=false, font=footnotesize]{subfig}
\usepackage{circuitikz}
\usepackage{bm}
\usepackage{stfloats}
\usepackage{microtype}
\usepackage{amsthm}
\usepackage{enumitem} 

\newcommand{\emaildot}{\makebox[0.2em]{\scalebox{.25}{\textbullet}}}
\newtheorem{thm}{Theorem}

\graphicspath{ {./images/} }

\begin{document}

\title{Implicit Neural Representation for Mesh-Free Inverse Obstacle Scattering\\
\thanks{Tin Vlašić acknowledges support from the Swiss Government through the Swiss Federal Commission for Scholarships for Foreign Students (FCS), and the Croatian Science Foundation under Project IP-2019-04-6703. Hieu Nguyen, AmirEhsan Khorashadizadeh, and Ivan Dokmanić were supported by the European Research Council Starting Grant 852821--SWING.}%
}

\author{
    \IEEEauthorblockN{Tin Vlašić\IEEEauthorrefmark{1}\IEEEauthorrefmark{2}, Hieu Nguyen\IEEEauthorrefmark{2}, AmirEhsan Khorashadizadeh\IEEEauthorrefmark{2}, Ivan Dokmanić\IEEEauthorrefmark{2}
    }
    \IEEEauthorblockA{\IEEEauthorrefmark{1}  \textit{Faculty of Electrical Engineering and Computing, University of Zagreb, Zagreb, Croatia}}
    \IEEEauthorblockA{\IEEEauthorrefmark{2} \textit{Department of Mathematics and Computer Science, University of Basel, Basel, Switzerland}\\
Email: \href{mailto:tin.vlasic@fer.hr}{tin.vlasic}@fer\emaildot hr, \{\href{mailto:hieuhuu.nguyen@unibas.ch}{hieuhuu.nguyen}, \href{mailto:amir.kh@unibas.ch}{amir.kh}, \href{mailto:ivan.dokmanic@unibas.ch}{ivan.dokmanic}\}@unibas\emaildot ch}
}

\maketitle

\begin{abstract}
Implicit representation of shapes as level sets of multilayer perceptrons has recently flourished in different shape analysis, compression, and reconstruction tasks. In this paper, we introduce an implicit neural representation-based framework for solving the inverse obstacle scattering problem in a mesh-free fashion. We express the obstacle shape as the zero-level set of a signed distance function which is implicitly determined by network parameters. To solve the direct scattering problem, we implement the implicit boundary integral method. It uses projections of the grid points in the tubular neighborhood onto the boundary to compute the PDE solution directly in the level-set framework. The proposed implicit representation conveniently handles the shape perturbation in the optimization process. To update the shape, we use PyTorch's automatic differentiation to backpropagate the loss function w.r.t. the network parameters, allowing us to avoid complex and error-prone manual derivation of the shape derivative. 
Additionally, we propose a deep generative model of implicit neural shape representations that can fit into the framework. The deep generative model effectively regularizes the inverse obstacle scattering problem, making it more tractable and robust, while yielding high-quality reconstruction results even in noise-corrupted setups.
\end{abstract}


\section{Introduction}
\begin{figure}[t]
    \centering
    \includegraphics[width=0.4572\columnwidth]{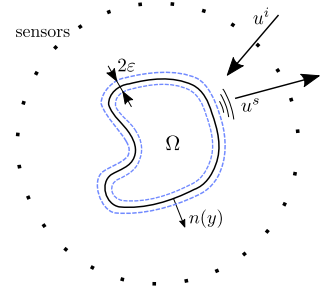}
    \caption{\textbf{Inverse obstacle scattering setup.} $u^i$ is the incident wave, and $u^s$ is the wave scattered by the obstacle (solid line). The sensors record the scattered waves at locations surrounding the obstacle. $n(y)$ is the normal vector at the point $y$. Dashed blue lines denote a thin tubular neighborhood around the obstacle boundary.}
    \vspace{-1em}
    \label{fig:helmholtzSetup}
\end{figure}

In the inverse scattering problem \cite{Kirsch2011AnIntroToInverse, colton2013integralEqMethods, Colton2018lookingback, colton2019inverse}, the scattering object is an obstacle with given boundary conditions, and the goal is to determine the obstacle from a set of measurements of the scattered field $u^s$. In this paper, we consider the sound-hard acoustic scattering problem (see Fig.~\ref{fig:helmholtzSetup}) whose forward model is given by the Helmholtz equation:
\begin{equation}
    \label{eq:pde}
    \begin{aligned}
        (\Delta + \kappa^2)u &= 0 ~\mathrm{in}~ \mathbb{R}^d \setminus \bar \Omega, \\
        \frac{\partial u}{\partial n} &= 0 ~\text{on}~ \partial \Omega,
    \end{aligned}
\end{equation}
where  $\Delta$ is the Laplacian operator, $\kappa$ is the wave number, $\Omega$ is the obstacle domain with boundary $\partial \Omega$, $n$ is the outward unit normal to the boundary, and ${u=u^i+u^s}$ for $u^i$ is an incident wave field. To ensure a unique forward solution, there is also the radiation condition imposed on the scattered wave:
\begin{equation}
    \lim_{r \rightarrow \infty} r^{d-1} \left( \frac{\partial u^s}{\partial r}  - j \kappa u^s \right) = 0 ~ \mathrm{for} ~ r=\Vert x \Vert.
\end{equation}
A method to solve the direct obstacle scattering problem is the integral equation method \cite{colton2013integralEqMethods, Kress1991, folland2020introductionpde}. By formulating the problem in terms of boundary integral equations, the method reduces the dimensionality of the problem and avoids the application of absorbing boundary conditions \cite{colton2013integralEqMethods}.

To solve the inverse scattering problem, we are interested in those methods that easily handle topological changes of the boundary. One such method is the implicit boundary integral method (IBIM) \cite{kublik2013implicit, chen2017implicit, izzo2021correctedImplicit} which approximates the boundary integral via volume integral whose domain is implicitly represented by the signed distance function (SDF). Such a representation is advantageous in that it handles complicated shapes and their moving boundaries easily without the need for costly frequent remeshing, making the inverse problem more tractable \cite{santosa_1996, Osher_2003, BURGER_2005}.

However, the grid-based representation of an SDF, especially in 3D space, is memory-inefficient. Recently, \textit{implicit neural representations} of shapes \cite{Park_2019_CVPR, chen2018implicit_decoder, atzmon2019controlling, icml2020_2086}, where a multilayer perceptron (MLP) ${\eta_{\theta}: \mathbb{R}^d \rightarrow \mathbb{R}}$ approximates the SDF, have shown remarkable results in data compression. Instead of being related to the fixed discretization of the ambient 3D space, the parameters $\theta$ of the MLP are directly related to the shape, whose watertight surface is implicitly represented by the isosurface of ${\eta_{\theta}(x)=0}$. Among others, SIREN \cite{sitzmann2019siren}, which is an MLP architecture with periodic activation functions, fits SDFs of complicated shapes robustly.

In this paper, we propose using an implicit neural representation of an obstacle shape in the inverse scattering problem. It allows us to represent various shapes with a number of parameters smaller than the number of grid points of interest. Contrary to the deep learning shape optimization frameworks, such as \cite{remelli2020meshsdf, deepMesh2021}, where the Marching Cubes algorithm \cite{Lorensen_1987, Sethian1999fmm} extracts the surface from the neural SDF on some predefined grid prior to solving the PDE, we use the closest point formulation proposed in \cite{kublik2013implicit} and \cite{chen2017implicit} for computing boundary integrals directly in the level-set framework. The loss functional is composed of a data-fidelity term and penalty terms on the SDF to regularize the ill-posedness of the inverse obstacle scattering problem. We implement the proposed framework in PyTorch \cite{pytorch2019}, making it differentiable throughout via automatic differentiation (AD). This allows for convenient gradient-based optimization of the parameters $\theta$, and avoids manual derivation of shape derivatives. While level-set methods are not new in the inverse scattering problems \cite{Dorn_2006, Guo_2013, Albuquerque_2021}, to the best of our knowledge, implicit neural representations of shapes were not yet taken advantage of in such problems. The continuous parameterization with a SIREN offers additional benefits over grid-based representations such as the analytical computation of gradients and higher-order derivatives that are independent of conventional grid resolutions. The experimental results demonstrate that the proposed framework achieves high-quality reconstruction results even in noise-corrupted setups.

Additionally, we demonstrate that deep generative models (DGMs) of implicit neural representations can be implemented in the proposed framework for task-adaptive inverse obstacle scattering, i.e., for the inverse scattering of particular objects that look alike. Recently, DGMs have shown effectiveness in regularizing inverse problems by constraining their solutions to remain on a learned manifold~\cite{bora2017csgm, ongie2020deep, daras2021solving}. Analogously, we propose a normalizing flow-based DGM of implicit neural representations for regularizing the solution of the inverse obstacle scattering problem to come from the same distribution as training data, i.e., that the reconstructed shape has similar features as the shapes from a training dataset. The proposed DGM utilizes real-valued non-volume preserving (real NVP)~\cite{dinh2016density} transformations for density estimation of low-dimensional feature vectors of the training data, making it stable and memory efficient to train. When the proposed DGM regularizes the inverse scattering problem, the optimization problem gets a simple expression consisting of only the data-fidelity term. The solution of the inverse problem is optimized over the low-dimensional input latent vector that has only a few degrees of freedom and leads to faster convergence. The space of possible solutions is rather small which makes the inverse problem even more tractable. 

\section{Direct Obstacle Scattering}
The solution of \eqref{eq:pde} can be represented by the single layer potential \cite{folland2020introductionpde}
\begin{eqnarray}
    \label{eq:singleLayerPotential}
    u^s(x) = \mathcal{S}[\alpha](x) = \int_{\partial \Omega} \alpha(y) \Phi (x,y) ds(y)  ~\mathrm{in}~ \mathbb{R}^d {\setminus} \partial \Omega,
\end{eqnarray}
where $\alpha$ is the density to match the boundary condition and $\Phi$ is Green's function. Green's function for the Helmholtz equation and ${d=2}$ is
\begin{equation}
    \Phi(x,y) = \frac{j}{4}H_0^1(\kappa\Vert x-y \Vert),
\end{equation}
where $H_0^1$ is the first kind Hankel function of degree $0$. The density ${\alpha \in C(\partial \Omega)}$ satisfies a boundary integral equation
\begin{eqnarray}
\label{eq:bie}
    \int_{\partial \Omega} \alpha(y) \dfrac{\partial \Phi (x,y)}{\partial n(x)} ds(y) - \dfrac{1}{2} \alpha(x) = -\frac{\partial u^i}{\partial n}(x) ~\text{on}~ \partial \Omega.
\end{eqnarray}
Finding a solution involves two steps: \textit{i)} solving the linear system in \eqref{eq:bie} for $\alpha(x)$ and \textit{ii)} evaluating \eqref{eq:singleLayerPotential}. A key challenge is to accurately evaluate the boundary integral in \eqref{eq:bie} where $\partial \Omega$ is represented implicitly as the zero-level set of the SDF. This problem has been addressed by the IBIM \cite{kublik2013implicit, chen2017implicit} where an effective solution involves a volume integration over the tubular neighborhood (please refer to Fig. \ref{fig:helmholtzSetup})
\begin{equation}
    T_{\varepsilon} = \{ x \in \mathbb{R}^d: \vert \eta_{\theta}(x) \vert \leq \varepsilon\}, ~\text{for}~ \varepsilon > 0,
\end{equation}
and a projection of the points in $T_{\varepsilon}$ onto the boundary given by:
\begin{equation}
    \label{eq:projection}
    P_{\partial \Omega}(x) = x - \eta_{\theta}(x)\nabla \eta_{\theta}(x), ~x \in T_\varepsilon,
\end{equation}
since the SDF satisfies the Eikonal equation $\vert \nabla \eta_{\theta}(x) \vert = 1$ for all $ x \in T_\varepsilon$  if $\partial \Omega$ is sufficiently smooth, and the normal $n(x) = - \nabla \eta_{\theta}(x)$ for $\{x\in \mathbb{R}^d : \eta_{\theta}(x)=0 \}.$
\begin{thm}[Kublik \textit{et al.} \cite{kublik2013implicit}] 
Assume a surface ${\partial \Omega \subset \mathbb{R}^d}$ is $C^2$ compact and $\eta_\theta$ is the signed distance function to the surface. For any function $f$ continuous on the surface, we have
\begin{equation}
    \int_{\partial \Omega} f(y) ds(y) = \int_{T_\epsilon} f(P_{\partial \Omega}(y)) J_{\partial \Omega}(y) \delta_\varepsilon(\eta_\theta(y))dy
\end{equation}
for any distributional kernel $\delta_\varepsilon$ compactly supported in ${[-\varepsilon, \varepsilon]}$, and the Jacobian $J_{\partial \Omega}$ which is in the level-set framework for ${d=2}$ given by
\begin{equation}
    J_{\partial \Omega}(x) = 1 - \eta_\theta(x)\Delta \eta_\theta(x).
\end{equation}
\end{thm}

Solving the boundary integral equation \eqref{eq:bie} directly in the level-set framework by using the IBIM effectively avoids grid-size-dependent extraction methods of surface points and quantization errors related to them. The extraction methods such as Marching Cubes are often non-differentiable and heuristic techniques have to be used to make them employable in the shape optimization tasks that require backpropagation. In contrast, the projection of the points onto the obstacle surface is differentiable and straightforward for backpropagation. Furthermore, the continuous parameterization of the SDF with a SIREN is convenient for solving the boundary integral equation since it leads to analytical computation of the expressions such as the normal derivative and the Jacobian $J_{\partial \Omega}$. Notice that, unlike the rectified linear unit (ReLU), the sine activation function has higher-order derivatives.

\begin{figure}[t]
    \centering
    \includegraphics[width=0.7\columnwidth]{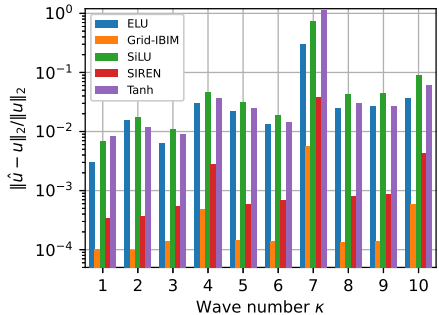}
    \vspace{-0.85em}
    \caption{\textbf{Relative $\ell_2$ errors of direct scattering solutions for different network architectures.} To fit the unit-circle SDF, we initialized all networks with $2$ hidden layers, $128$ hidden features, and trained for $5000$ iterations. The graph shows medians of relative $\ell_2$ errors for $25$ direct problem simulations.}
    \vspace{-1em}
    \label{fig:directProblem}
\end{figure}

We compared different neural network architectures with different activation functions, namely SIREN, Tanh, SiLU, and ELU, in solving a direct scattering problem adopted from~\cite{chen2017implicit}. The problem consisted of a unit circle, whose SDF is given analytically as ${sdf(x) = 1-\Vert x \Vert_2}$, and an incident wave for which there is an analytical solution. The analytic SDF was evaluated on a ${128 \times 128}$ computational grid of a $[-1.275, 1.275]^2$ domain. We evaluated the solution on a ${64 \times 64}$ uniform grid of a domain ${[-5, 5]^2 \setminus \Omega}$ and calculated the relative error to the analytical solution. The results in Fig.~\ref{fig:directProblem} show that the framework yields the smallest relative errors when the SIREN architecture is used. They also show that for the SIREN architecture, the direct solution errors are slightly worse than for the grid-based IBIM but comparable, which is to be expected in this example since in our framework the SDF values are approximated with an MLP and in the grid-based IBIM the SDF values are given analytically. Based on these results, the SIREN architecture was employed in the framework for solving the inverse obstacle scattering problem.

\begin{figure*}[b]
    \centering
    \includegraphics[width=0.8\textwidth]{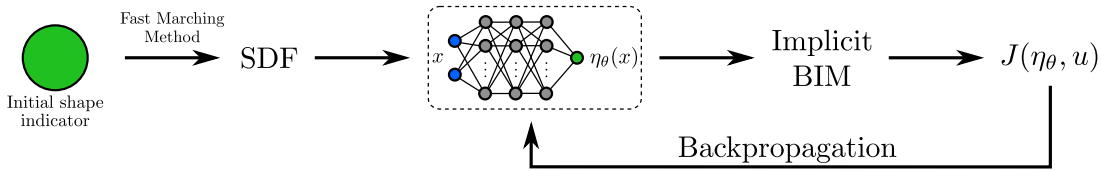}
    \caption{\textbf{Shape reconstruction pipeline.} The MLP is first trained to represent an initial shape before being employed in the inverse problem. By using the neural SDF, we solve the direct problem to obtain the predicted scattered field. Minimization of the loss function updates network parameters, i.e., the shape.}
    \label{fig:proposedFramework}
\end{figure*}

\section{Physics-Based Inverse Obstacle Scattering}
\label{sec:inverseObstScatt}
Representing a shape through the SDF has several advantages in the shape reconstruction problems \cite{santosa_1996}:
\begin{itemize}[noitemsep, nolistsep]
    \item No a priori assumptions on the topology of the shape $\Omega$ need to be made. For example, the shape could be made of several disconnected subregions.
    \item No a priori assumptions on the nature of the shape $\Omega$ need to be made. In the scattering literature \cite{colton2019inverse}, there is often an assumption that the unknown obstacle is star-shaped.
    \item It conveniently handles complicated shapes and their perturbations, i.e., moving boundaries, without the need of constant remeshing between the iterations of the optimization process.
\end{itemize}
These advantages in addition to the level-set framework-based IBIM make the shape representation through the SDF an ideal fit for solving the inverse obstacle scattering problem.

The most crucial component in a shape reconstruction problem is the shape derivative. For demanding shape reconstruction and optimization problems, derivation of the shape derivative is often complex and error-prone, leading to the emergence of AD~\cite{eppler2019computation, dokken2020automatic}. We implement the implicit neural shape representation and the IBIM in PyTorch, making the forward model differentiable throughout via PyTorch's AD engine.

Given $M$ measurements $u(x_m)$ associated with sensor locations ${x_m \in \mathbb{R}^d {\setminus} \Omega}$, the goal of the inverse obstacle scattering problem is to reconstruct the obstacle shape for which the calculated solution $\hat{u}(x_m)$ of the boundary value problem \eqref{eq:pde} is the closest possible to the measurements. Since in the proposed framework the obstacle shape is determined by the network parameters $\theta$, the problem can be formulated as:
\begin{equation}
    \label{eq:lossFunc}
    \min_{\theta} \mathcal{J}(\eta_\theta, u) = \frac{1}{M} \sum_{m=1}^{M} \vert u(x_m) - \hat{u}(x_m) \vert^2 + \mathcal{R}(\eta_\theta),
\end{equation}
where the regularizer is
\begin{equation}
    \label{eq:reg}
    \mathcal{R}(\eta_\theta) = \lambda_1 \sum_{n=1}^{N} \vert \vert \nabla \eta_\theta(x_n) \vert - 1 \vert^2 + \lambda_2 \sum_{n=1}^{N} \vert \Delta \eta_\theta(x_n) \vert^2,
\end{equation}
for $\{x_n\}$ are points in a region of interest (ROI). The ROI is selected to cover the obstacle and sampled in a multiresolution way such that the majority of the $\{x_n\}$ points are close to the zero-level set of the SDF. The first term in \eqref{eq:reg} ensures that $\eta_\theta$ is an SDF and the second term imposes its smoothness.

The inverse obstacle scattering pipeline is depicted in Fig~\ref{fig:proposedFramework}. Since the forward model is differentiable, we can employ PyTorch's backpropagation algorithm to calculate the gradients of the loss function w.r.t. the parameters $\theta$ and update the parameters accordingly within the gradient descent procedure.

\section{Data-Driven Inverse Obstacle Scattering}
\label{sec:dgmScattering}
We extend the proposed framework to work with DGMs of implicit neural shape representations. A DGM can be trained to output samples that come from the same distribution as a training dataset of SDFs. The purpose of the DGM is then to regularize the task-adaptive inverse obstacle scattering problem by constraining the reconstructed shape to remain on a learned manifold. The DGM is trained on a specific dataset of SDFs (e.g., SDFs of car shapes) so that its samples approximately satisfy the Eikonal equation and their zero-level sets are similar to the zero-level sets of the training SDFs.

\begin{figure}[t]
    \centering
    \includegraphics[width=0.85\columnwidth]{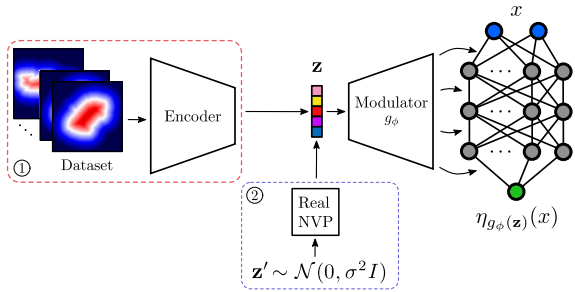}
    \caption{\textbf{Generative modeling of implicit neural representations.} 1) An encoder extracts features from a training dataset of SDFs and represents shapes as low-dimensional latent vectors $\{\mathbf{z}_k\}_{k=1}^K$. A latent vector $\mathbf{z}_k$ is mapped through a hypernetwork (modulator) $g_\phi$ to obtain the weights of a coordinate-based MLP $\eta_\theta$ that then represents the $k^{th}$ training shape. 2) Once the encoder and modulator are learned to faithfully represent various shapes from the dataset, a normalizing flow (real NVP) $f_\psi$ is trained to generate latent vectors ${\tilde{\mathbf{z}}=f^{-1}_\psi(\mathbf{z'})}$ that come from the same distribution as a set of the training latent vectors $\mathbf{z}_1, \mathbf{z}_2, \dots, \mathbf{z}_K$.}
    \vspace{-1em}
    \label{fig:gmScheme}
\end{figure}

\begin{figure}[t]
	\centering
		\subfloat[Training samples]{%
        	\includegraphics[width=0.485\columnwidth]{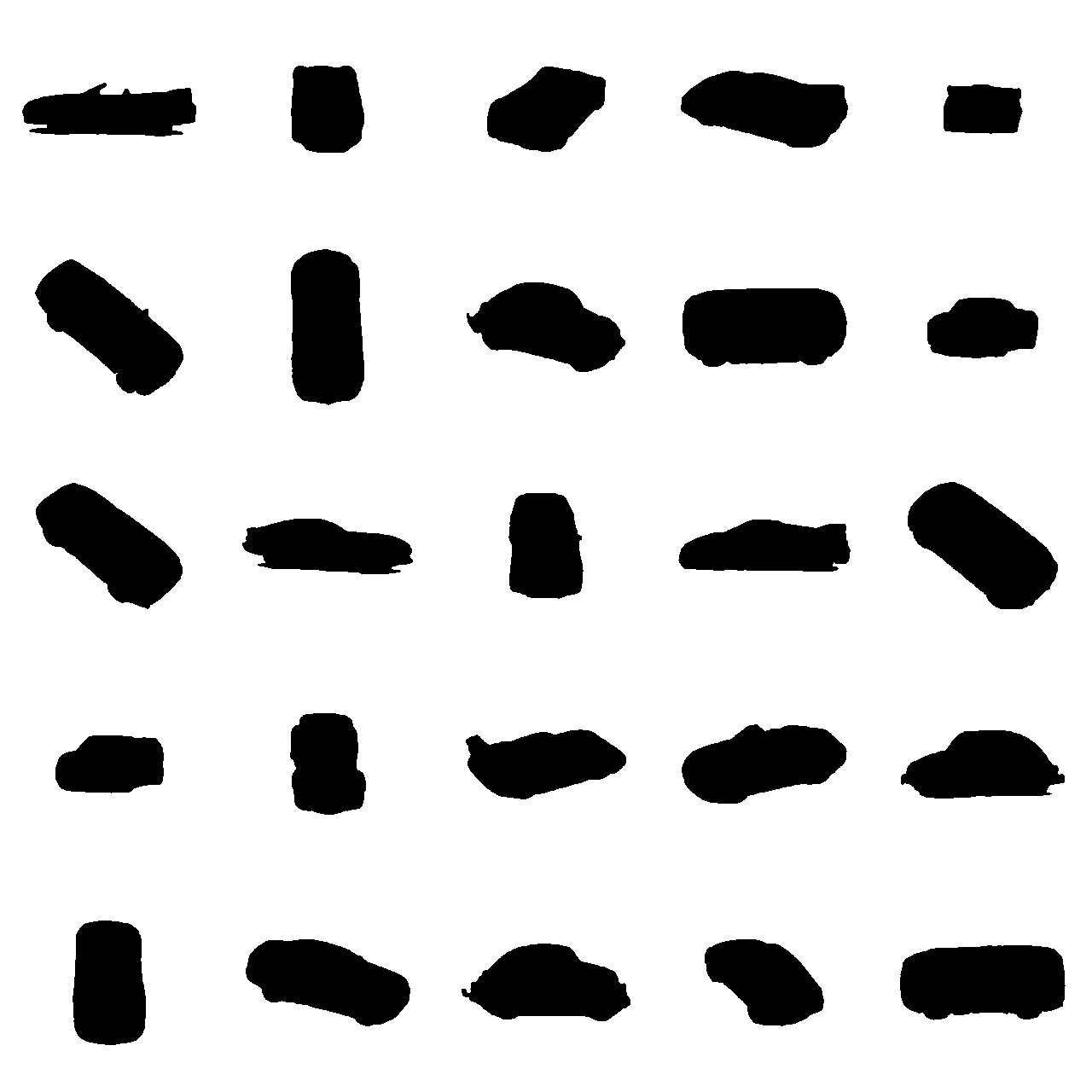}%
        }
        \hspace{0.25ex}
		\subfloat[Generated samples]{%
			\includegraphics[width=0.485\columnwidth]{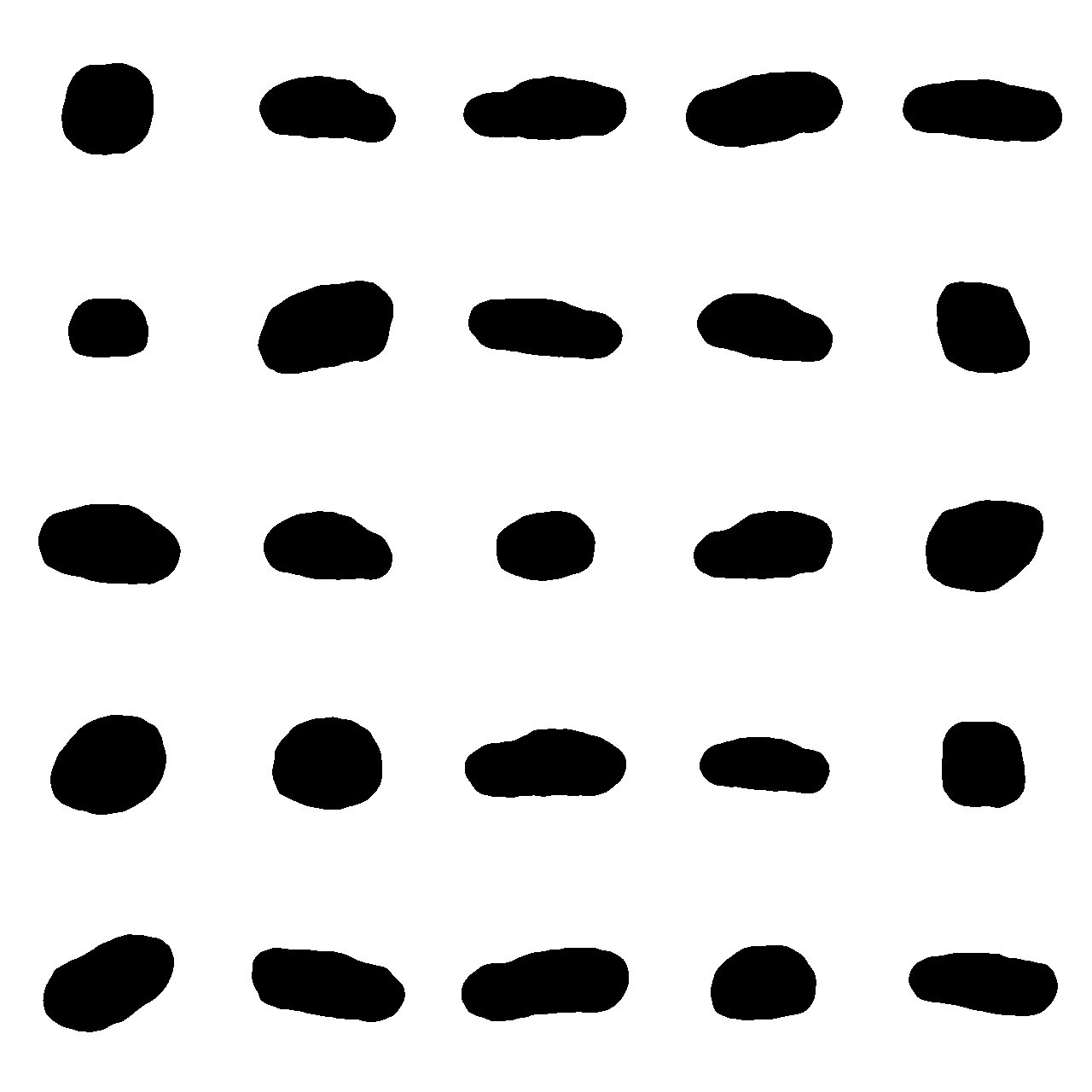}%
        }
	\caption{\textbf{Samples from our model trained on a dataset of car obstacle SDFs.} Each image in (b) corresponds to a continuous SDF evaluated on a ${256 \times 256}$ grid. Due to the presentation reasons, the samples are shown in the binary format where the positive values of the SDFs are mapped to black and negative to white color.}
	\vspace{-1em}
	\label{fig:generatedSamples}
\end{figure}

In this paper, we propose a normalizing flow-based DGM of implicit neural shape representations depicted in Fig. \ref{fig:gmScheme}.
The proposed DGM is learned in two steps: \textit{i)} an encoder is learned to extract features $\{\mathbf{z}_k\}$ from the input SDFs that are then mapped through a simultaneously trained hypernetwork $g_\phi$ to the weights $\theta$ of the SIREN MLP $\eta_\theta$; \textit{ii)} real NVP learns the distribution of the latent vectors $\{\mathbf{z}_k\}$ that correspond to the SDFs from the training dataset. Step \textit{i)} is obtained by minimizing the mean squared error (MSE) between the SDFs in the training dataset and the output of the coordinate-based MLP $\eta_{g_\phi(\mathbf{z})}$ at batch points sampled uniformly at random. Additionally, in the training procedure, the output is constraint such that the norm of its spatial gradients $\Vert \nabla \eta_{g_\phi(\mathbf{z})} \Vert$ is as close as possible to $1$ at the batch points. Once the encoder and modulator $g_\phi$ are trained, the SDFs from the training dataset are passed through the encoder to obtain the latent codes $\{\mathbf{z}_k\}$. Real NVP then learns a stable and invertible mapping $f_\psi$ between the distribution of the codes $\{\mathbf{z}_k\}$ and a Gaussian distribution. The inverse function $f^{-1}_\psi$ maps samples from the Gaussian distribution into approximate samples from the distribution of the latent codes, which corresponds to the generation of samples from the model. A generated latent code ${\tilde{\mathbf{z}}=f^{-1}_\psi(\mathbf{z'})}$ is then mapped through the hypernetwork $g_\phi$ to obtain the weights ${\theta = g_\phi(\tilde{\mathbf{z}})}$ of the coordinate-based MLP $\eta_\theta$, which consequently yields a new obstacle shape similar to the shapes from the training dataset. A subset of samples from a training dataset and the generated samples obtained by the proposed DGM are illustrated in Fig. \ref{fig:generatedSamples}.

Generative adversarial networks (GANs) \cite{goodfellow2014generative} can produce high-quality samples, but are often hard to train since the training procedure is unstable and they suffer from pathologies such as mode collapse \cite{thanh2020catastrophic}. Moreover, they are generally non-invertible, or computing the inverse is slow, thus they are not a good fit for downstream inference tasks. In contrast, normalizing flows are invertible and can be trained using maximum likelihood, making the training procedure stable. During inference time, they provide direct access to likelihoods which is often a desirable property. However, for high-dimensional probability distributions, normalizing flows are computationally intensive because the dimension of the input vector space equals the dimension of the generated samples. In this paper, the normalizing flow learns a distribution of the low-dimensional feature vectors  $\{\mathbf{z}_k\}$, leading to a stable, invertible, memory-efficient and computationally lightweight DGM.

The proposed DGM can be employed in the inverse obstacle scattering framework from Section \ref{sec:inverseObstScatt}. In this setting, the coordinate-based MLP in Fig. \ref{fig:proposedFramework}, which is first trained on a single initial SDF and whose weights $\theta$ are updated every iteration, is replaced by the DGM. The DGM is initialized by a low-dimensional sample ${\mathbf{z}' \sim \mathcal{N}(0, \sigma^2I)}$. The extensive experiments demonstrated that the mean Gaussian is usually a good starting point. Since the DGM constraints the solution to lie on a learned manifold, the loss function \eqref{eq:lossFunc} can be reduced to just the data-fidelity term. The error of the loss function is backpropagated all the way to the low-dimensional input vector $\mathbf{z}'$  which is updated every iteration to generate a shape that better fits the measurements. The low dimensionality of the input vector narrows the space of possible solutions and leads to the faster convergence of the inverse problem.

\section{Numerical Experiments}
We present the results of numerical simulations for inverse obstacle scattering in 2D. First, we give the reconstruction results for the physics-based framework proposed in Section~\ref{sec:inverseObstScatt} and then the results obtained by the data-driven framework proposed in Section~\ref{sec:dgmScattering}. For both frameworks, the simulations were conducted on objects from the ETH-80 dataset \cite{leibe2003analyzing}. The dataset consists of $10$ object categories and $410$ images of ${256 \times 256}$ size for each object. The ground truth SDFs were obtained on a ${256 \times 256}$ uniform grid defined on a ${[-1, 1]^2}$ square by the fast marching method (FMM) \cite{sethian1996fast}. 

We measured the scattered field at $M$ angular locations generated by scattering of multiple incident plane waves ${u^i_l(x)=e^{j\kappa \mathbf{x d}_l}}$ and multiple wave numbers $\kappa$. In the provided experiments, we uniformly displaced ${M=96}$ sensor locations on a circle of radius $4.5$ and the number of incident plane waves $u^i_l(x)$ was set to ${5}$ with directions $\mathbf{d}_l$ uniformly displaced in the space. We added white Gaussian noise to some of the measurements and express the quality of the measurement data in terms of the signal-to-noise ratio (SNR). Finally, the reconstruction quality was assessed in terms of Chamfer distance between the ground truth point set $S_{GT}$ and the reconstruction point set $S_{R,\kappa}$ for different $\kappa$ as:
\begin{multline}
    \label{eq:chamferDist}
    d_{C,\kappa}(S_{GT}, S_{R,\kappa}) = \frac{1}{2} \Biggl( \sum_{x \in S_{GT}} \min_{y \in S_{R,\kappa}}\Vert x-y \Vert^2 \\
     + \sum_{y \in S_{R,\kappa}} \min_{x \in S_{GT}}\Vert y-x \Vert^2 \Biggr).
\end{multline}

\begin{figure}[t]
	\centering
		\subfloat[${d_{C,7}{=}0.189}$, ${d_{C,14}{=}0.033}$]{%
        	\includegraphics[width=0.23\textwidth]{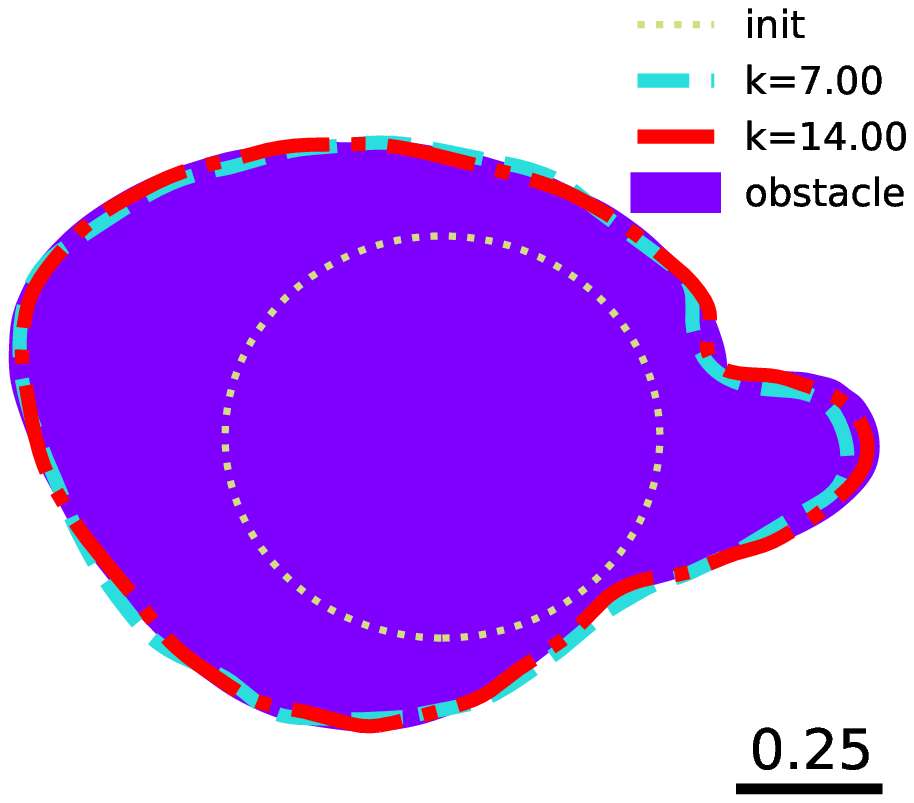}%
        }
        \hspace{0.5ex}
		\subfloat[${d_{C,7}{=}0.422}$, ${d_{C,14}{=}0.061}$]{%
			\includegraphics[width=0.23\textwidth]{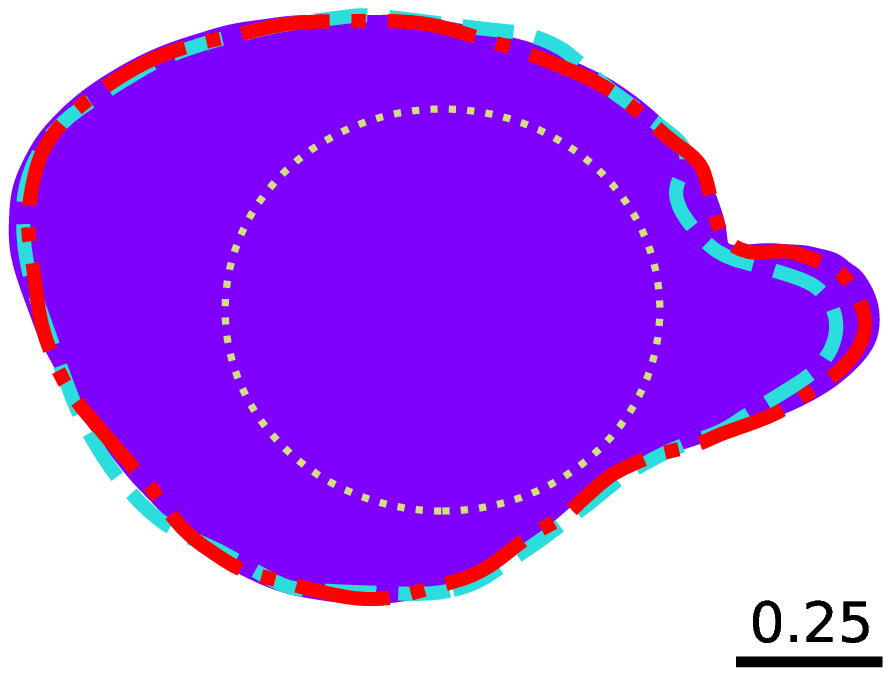}%
        }
        \vspace{-2.2ex}
        
		\subfloat[${d_{C,7}{=}1.528}$, ${d_{C,14}{=}0.145}$]{%
			\includegraphics[width=0.23\textwidth]{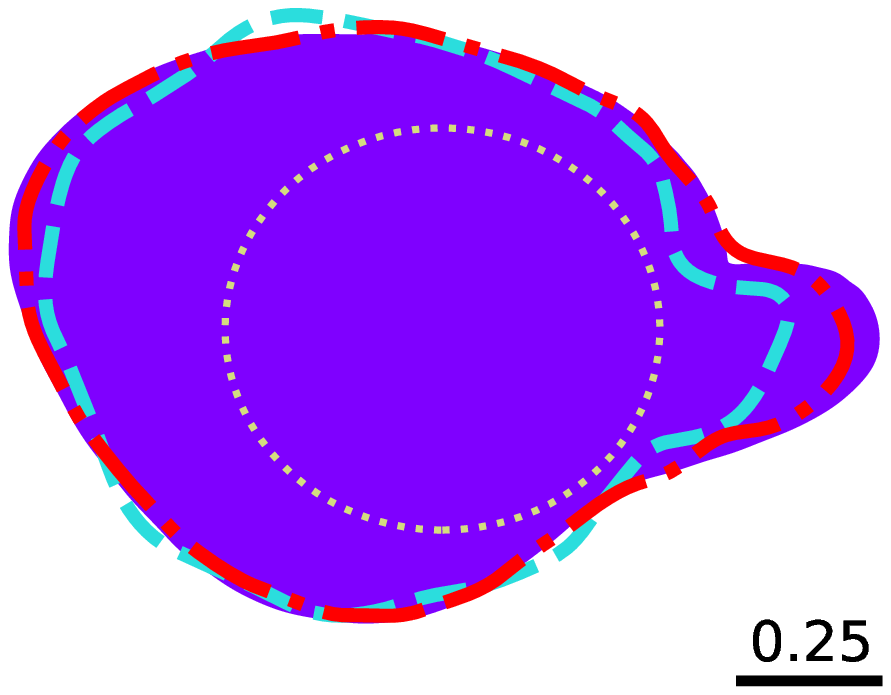}%
        }
        \hspace{0.5ex}
		\subfloat[${d_{C,7}{=}0.630}$, ${d_{C,14}{=}0.260}$]{%
			\includegraphics[width=0.23\textwidth]{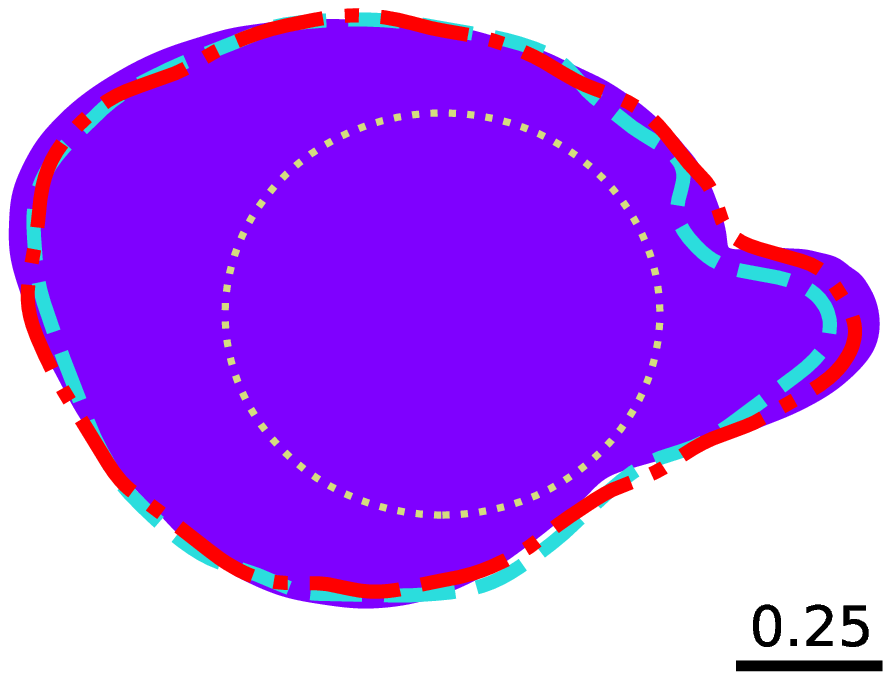}%
        }
	\caption{\textbf{Physics-based reconstruction of a 2D cup obstacle.} (a) Full-view noiseless; (b) Full-view noisy (${\text{SNR}=20}$ dB); (c) Limited-view noiseless; (d) Limited-view noisy measurements (${\text{SNR}=25}$ dB). In limited-view setups, sensors were uniformly displaced on intervals $[30^{\circ}, 150^{\circ}]$ and $[210^{\circ}, 330^{\circ}]$.}
	\vspace{-1em}
	\label{fig:rec_cup2}
\end{figure}

\subsection{Results Obtained using Physics-Based Framework}
In the experiments for the physics-based framework, we initialized the SIREN with $2$ hidden layers and $128$ hidden features, making it $33.5$k parameters in total, and we trained the network using PyTorch and Adam \cite{KingmaB14adam}. The ROI was set as a $[-1.1, 1.1]$ square. The initial shape was a circle of radius $0.4$ whose SDF was evaluated on a ${256 \times 256}$ uniform grid by the FMM. The obtained SDF was used to train the initial weights $\theta$ of the SIREN. 

The initialized SIREN was then employed in solving the IBIM and subsequently in solving the inverse problem iteratively. To solve the loss function \eqref{eq:lossFunc}, the ROI was sampled in a multiresolution way, with a few points outside the tubular neighborhood to control the SDF, and the majority of them inside for accurately solving the boundary integral equation. We report that the framework is very sensitive to the regularization parameters $\lambda_1$ and $\lambda_2$ which have to be chosen with care. For every wave number $\kappa_i$, we trained the network for $8000$ iterations with a learning rate $10^{-6}$. In the reconstruction process, the steps between different wave numbers $\kappa_i$ were chosen to be $0.5$, with the starting ${\kappa_0=1.5}$. For all ${\kappa_i > 1.5}$, the starting point of the minimization problem is the last point of the previously finished minimization problem for $\kappa_{i-1}$.

\begin{figure}[t]
	\centering
		\subfloat[${d_{C,8}{=}0.254}$, ${d_{C,15}{=}0.063}$]{%
        	\includegraphics[width=0.23\textwidth]{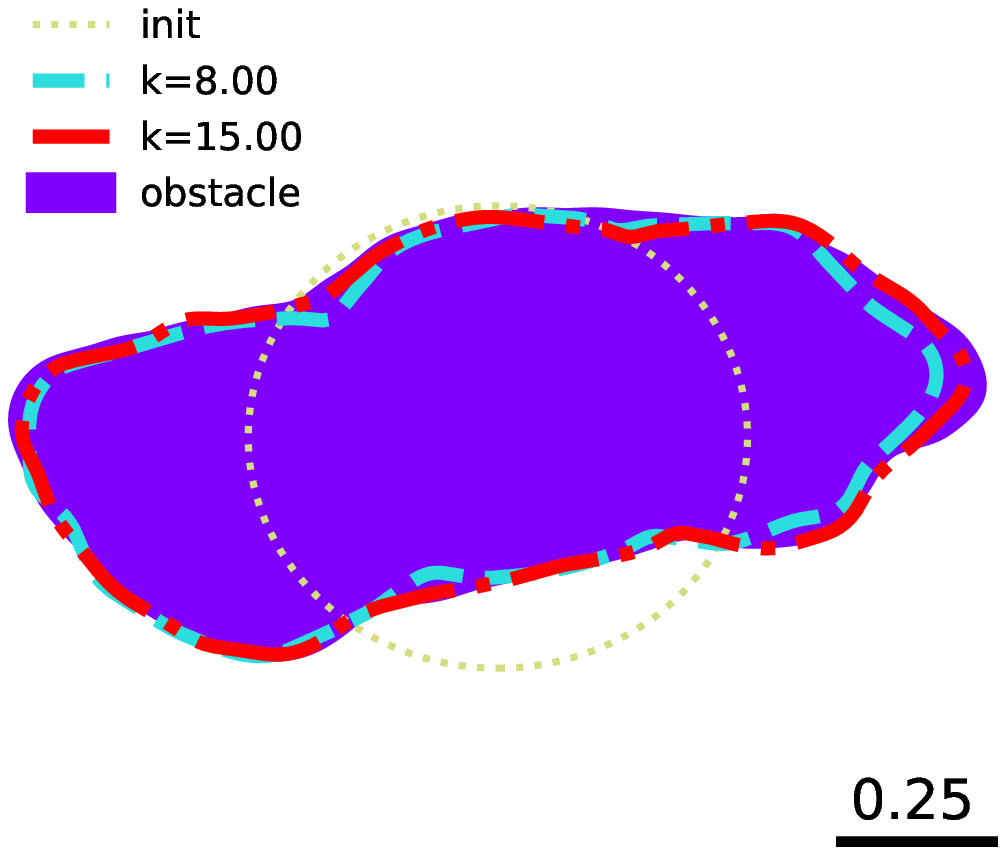}%
        }
        \hspace{0.5ex}
		\subfloat[${d_{C,8}{=}0.754}$, ${d_{C,15}{=}0.107}$]{%
			\includegraphics[width=0.23\textwidth]{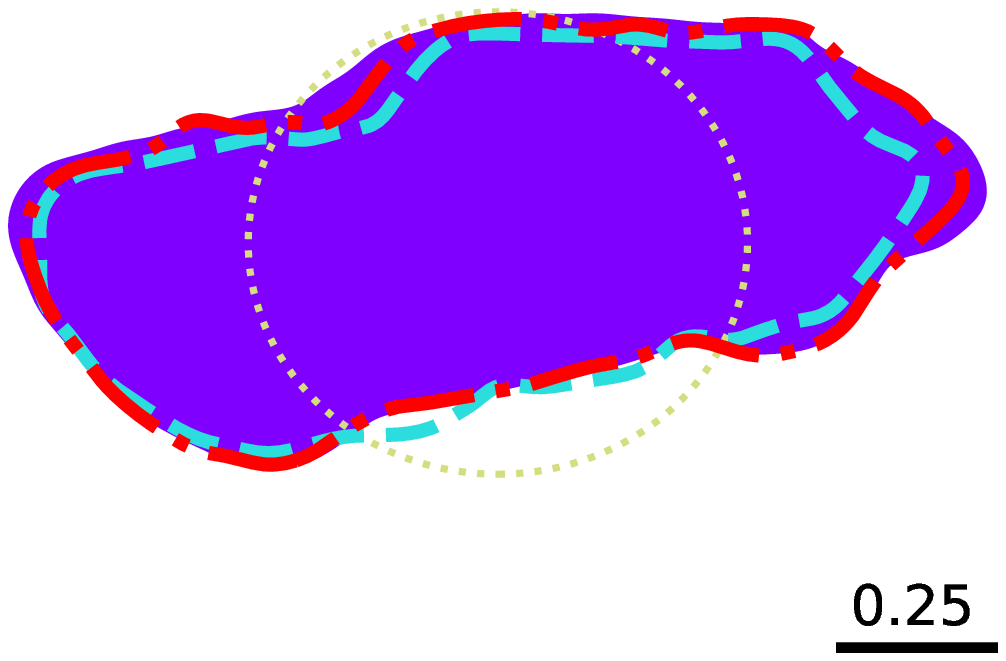}%
        }
        \vspace{-2.2ex}
        
		\subfloat[${d_{C,8}{=}0.959}$, ${d_{C,15}{=}0.126}$]{%
			\includegraphics[width=0.23\textwidth]{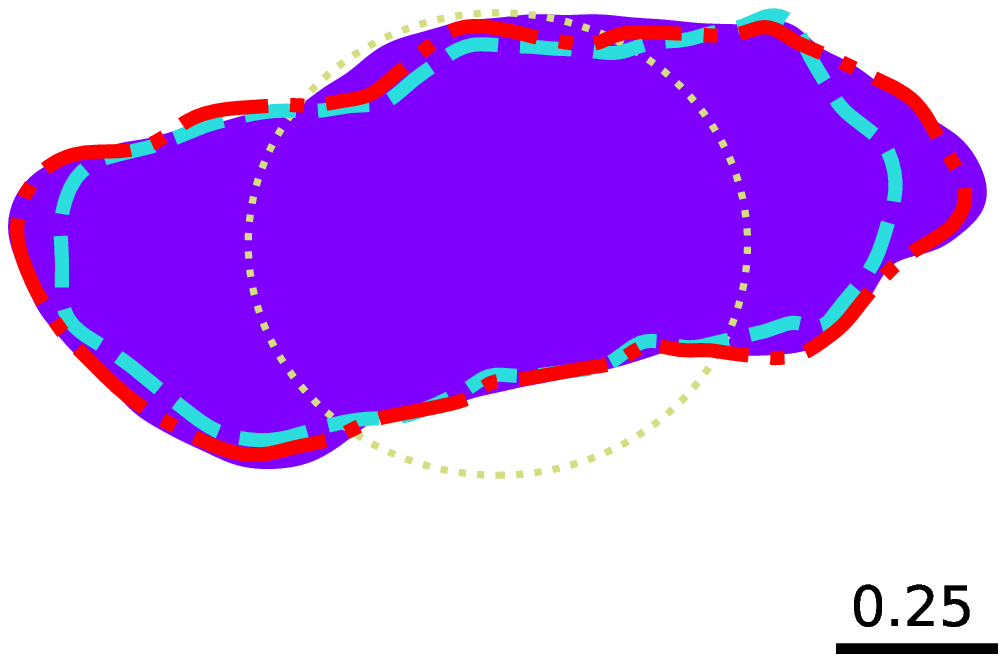}%
        }
        \hspace{0.5ex}
		\subfloat[${d_{C,8}{=}0.882}$, ${d_{C,15}{=}0.376}$]{%
			\includegraphics[width=0.23\textwidth]{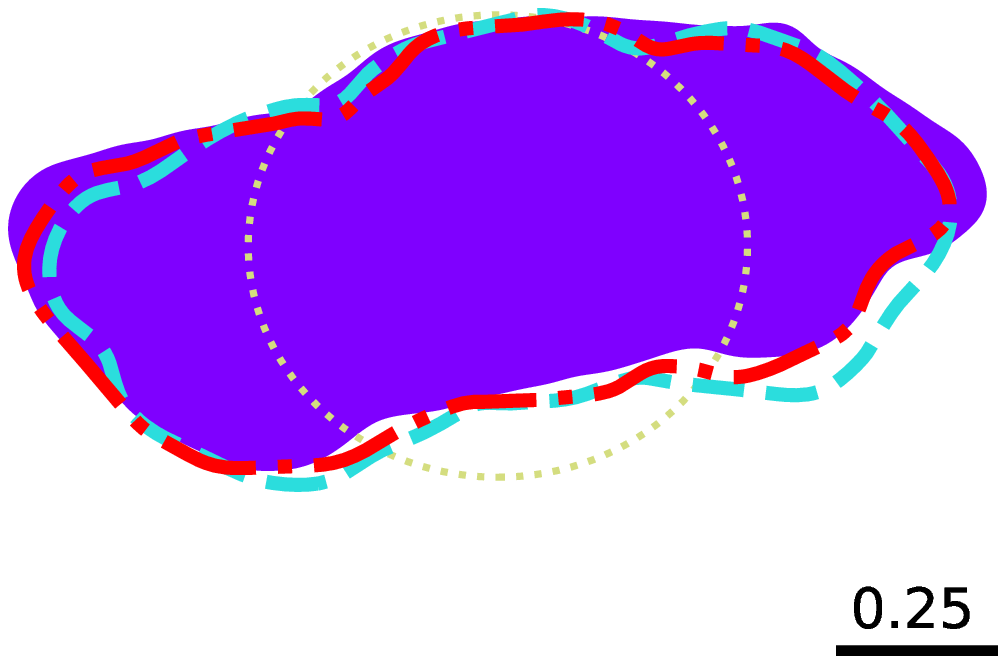}%
        }
	\caption{\textbf{Physics-based reconstruction of a 2D car obstacle.} (a) Full-view noiseless; (b) Full-view noisy (${\text{SNR}=25}$ dB); (c) Limited-view noiseless; (d) Limited-view noisy measurements (${\text{SNR}=30}$ dB). In limited-view setups, sensors were uniformly displaced on intervals $[-60^{\circ}, 60^{\circ}]$ and $[120^{\circ}, 240^{\circ}]$.}
	\vspace{-1em}
	\label{fig:rec_car1}
\end{figure}

We show some of the reconstruction results for cup and car objects in Fig. \ref{fig:rec_cup2} and \ref{fig:rec_car1}. The results demonstrate the ability of the framework to achieve high-quality reconstructions and its robustness in noise-corrupted and limited-view setups.

\subsection{Results Obtained using Data-Driven-Based Framework}
In the data-driven framework, we initialized the SIREN in the DGM with $5$ hidden layers and $256$ hidden features. The dimension of the latent code $\mathbf{z}$ was set to $32$. We learned the DGM on a dataset of $375$ car SDFs with $35$ shapes left for testing. The ROI was set as a $[-1,1]^2$ square. The DGM was initialized with the mean Gaussian.

We trained the network for 6000 iterations with a learning rate set to $10^{-2}$. Here, the minimization problems for $10$ different $\kappa$ values (from $1.5$ to $15$ with the step of $1.5$) were trained in parallel. The DGM-based inverse scattering setup is much easier and more robust to train since it is regularized by the generative model and the loss function has only one term. The data-driven model converges much faster than the physics-based model, achieving a faithful reconstruction after just a few hundred iterations. At the end of the training process, the parameters of the coordinate-based MLP can be fine-tuned to achieve more accurate reconstructions if needed.

\begin{figure}[t]
	\centering
		\subfloat[${d_{C,15}{=}0.318}$, ${d_{C,fine}{=}0.066}$]{%
        	\includegraphics[width=0.23\textwidth]{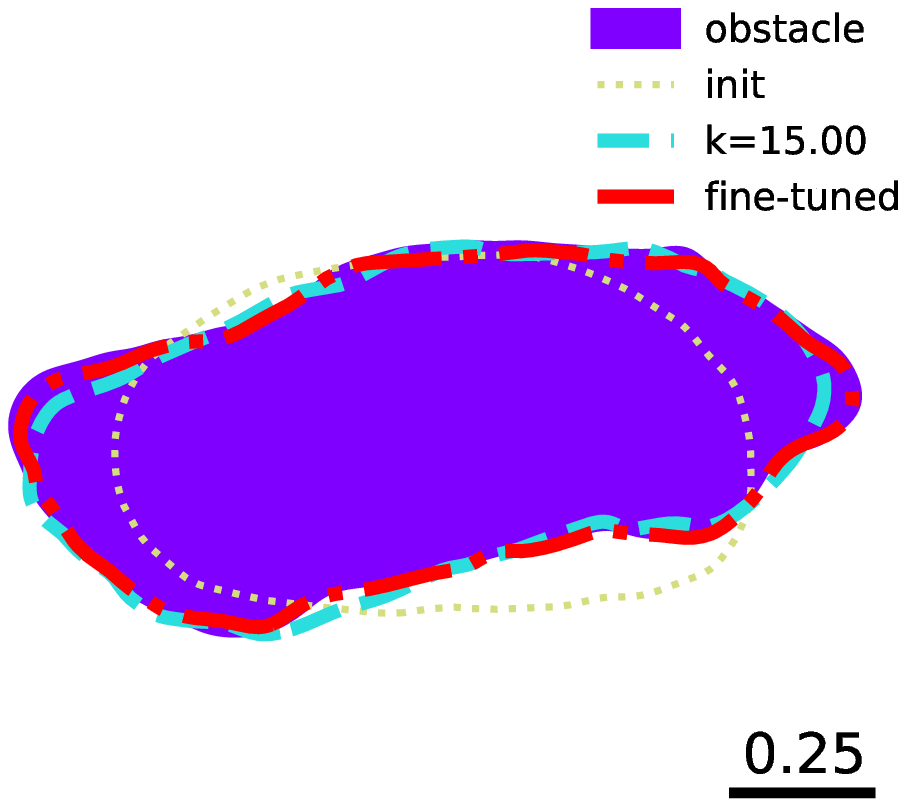}%
        }
        \hspace{0.5ex}
		\subfloat[${d_{C,15}{=}0.272}$, ${d_{C,\mathrm{fine}}{=}0.057}$]{%
			\includegraphics[width=0.23\textwidth]{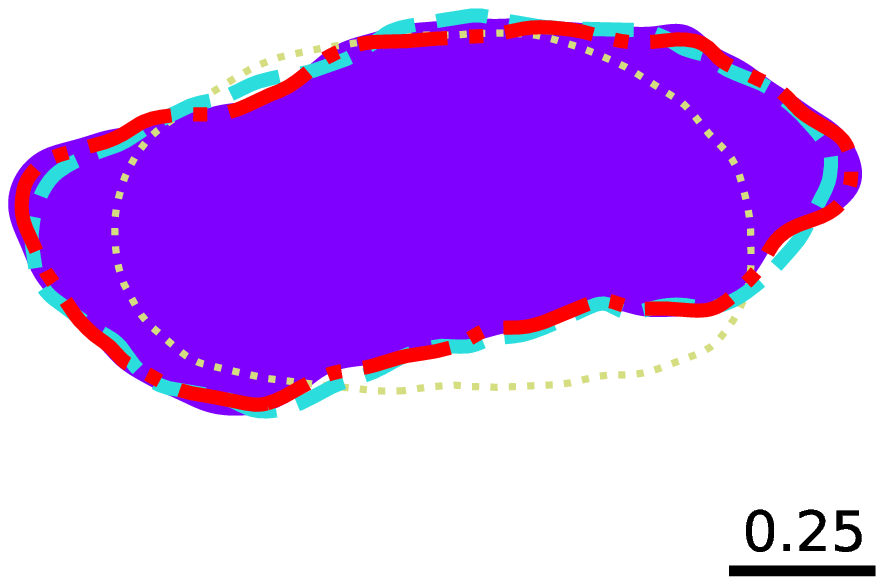}%
        }
        \vspace{-2.2ex}
        
		\subfloat[${d_{C,15}{=}0.454}$, ${d_{C,fine}{=}0.273}$]{%
			\includegraphics[width=0.23\textwidth]{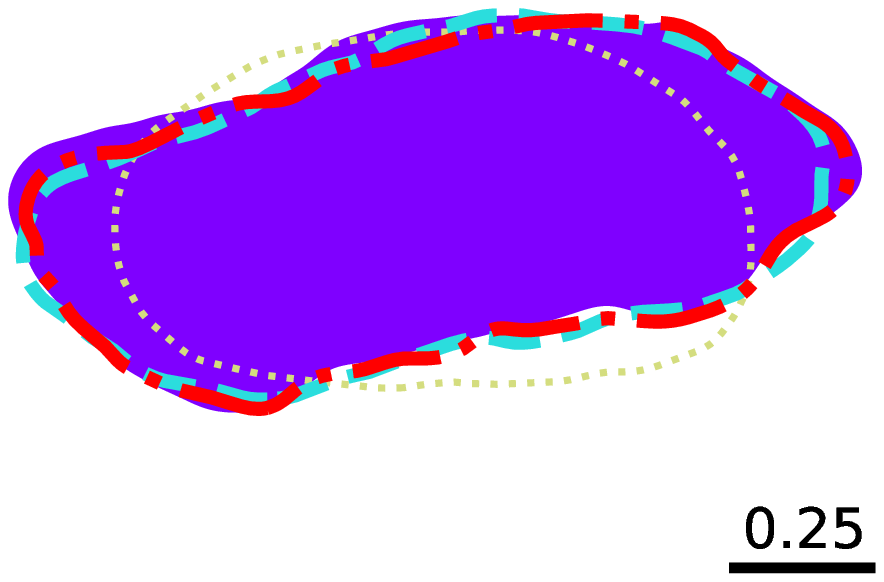}%
        }
        \hspace{0.5ex}
		\subfloat[${d_{C,15}{=}0.360}$, ${d_{C,fine}{=}0.171}$]{%
			\includegraphics[width=0.23\textwidth]{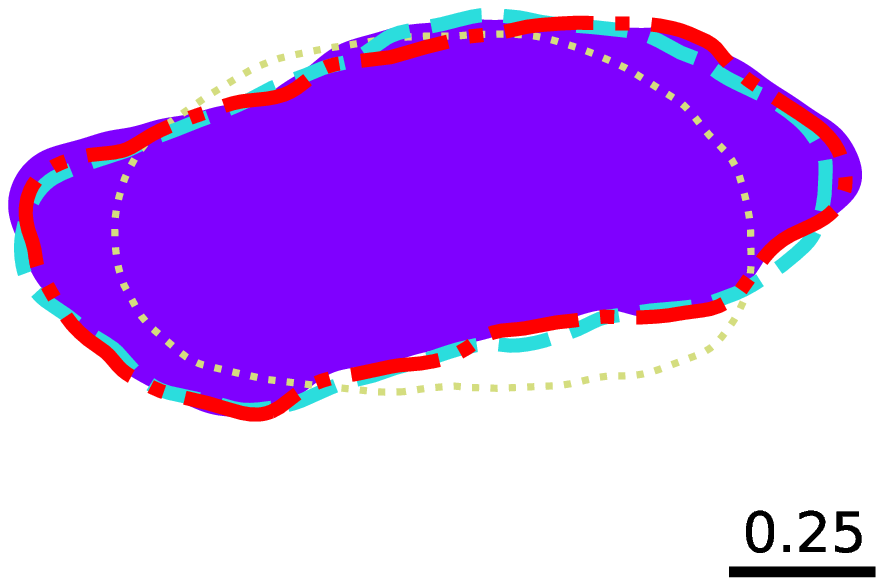}%
        }
	\caption{\textbf{Data-driven-based reconstruction of a 2D car obstacle.} (a) Full-view noiseless; (b) Full-view noisy (${\text{SNR}=25}$ dB); (c) Limited-view noiseless; (d) Limited-view noisy measurements (${\text{SNR}=30}$ dB). In limited-view setups, sensors were uniformly displaced on intervals $[-60^{\circ}, 60^{\circ}]$ and $[120^{\circ}, 240^{\circ}]$.}
	\vspace{-1em}
	\label{fig:rec_car1_dgm}
\end{figure}

We show some of the reconstruction results for a car obstacle in Fig. \ref{fig:rec_car1_dgm}. The results demonstrate the effectiveness and robustness of the proposed DGM in regularizing and solving the task-adaptive inverse obstacle scattering problem, including noisy and limited-view scenarios.

\section{Conclusion}
We proposed a novel implicit neural representation-based framework for solving the inverse obstacle scattering problem. Representing the obstacle shape with an MLP and solving the boundary value problem directly in the level-set framework leads to an accurate and robust mesh-free approach. We extended the proposed framework to work with a deep generative model which demonstrated the effectiveness in regularizing the task-adaptive inverse obstacle scattering problem. The proposed differentiable programming framework is user-friendly in that it avoids manual derivation of complex shape derivatives, easily handles complicated shapes, and achieves high-quality reconstruction results.

\balance

\bibliographystyle{myIEEEtran}
\bibliography{bibliography}

\end{document}